\documentclass[10pt, twocolumn, a4paper, table]{article}

\usepackage[table]{xcolor}
\usepackage{cvpr}
\usepackage{times}
\usepackage{epsfig}
\usepackage{graphicx}
\usepackage{amsmath}
\usepackage{amssymb}
\usepackage{color}
\usepackage{placeins}
\usepackage{multirow}
\usepackage{csquotes}
\usepackage{tabularx}
\usepackage[skip=2pt, justification=centering]{subcaption}
\usepackage[skip=5pt, font=small]{caption}
\usepackage{makecell}

\usepackage[pagebackref=true,breaklinks=true,colorlinks,bookmarks=false]{hyperref}
\usepackage[colorinlistoftodos,prependcaption]{todonotes}

\usepackage{wrapfig}
\usepackage{pdfpages}
\usepackage{sidecap}
\usepackage{colortbl}
\usepackage{overpic}
\usepackage{booktabs}
\usepackage{colortbl}
\usepackage[toc,page]{appendix}
\usepackage{placeins}
\usepackage{stfloats}
\usepackage{bookmark}

\usepackage{enumitem}

\usepackage{rotating}
\usepackage[export]{adjustbox}

\usepackage{xspace}

\usepackage{pifont}%

\usepackage[utf8]{inputenc}

\title{Improving Self-Supervised Single View Depth Estimation by Masking Occlusion}
\author{Maarten Schellevis\\Radboud University}

\cvprfinalcopy

\begin{document}

\maketitle

\begin{abstract}

Single view depth estimation models can be trained from video footage using a self-supervised end-to-end approach with view synthesis as the supervisory signal. 
This is achieved with a framework that predicts depth and camera motion, with a loss based on reconstructing a target video frame from temporally adjacent frames.
In this context, occlusion relates to parts of a scene that can be observed in the target frame but not in a frame used for image reconstruction.
Since the image reconstruction is based on sampling from the adjacent frame, and occluded areas by definition cannot be sampled, reconstructed occluded areas corrupt to the supervisory signal.
In previous work \cite{godard_digging_2018} occlusion is handled based on reconstruction error; at each pixel location, only the reconstruction with the lowest error is included in the loss. 
The current study aims to determine whether performance improvements of depth estimation models can be gained by during training only ignoring those regions that are affected by occlusion.

%
In this work we introduce occlusion mask, a mask that during training can be used to specifically ignore regions that cannot be reconstructed due to occlusions.
 Occlusion mask is based entirely on predicted depth information. 
 We introduce two novel loss formulations which incorporate the occlusion mask.
 The method and implementation of \cite{godard_digging_2018}  serves as the foundation for our modifications as well as the baseline in our experiments.
We demonstrate that (i) incorporating occlusion mask in the loss function improves the performance of single image depth prediction models on the KITTI benchmark. (ii) loss functions that select from reconstructions based on error are able to ignore some of the reprojection error caused by object motion.
    
\end{abstract}

\section{Introduction}
Knowing the depth of your surroundings is essential for navigating within an environment. We perceive the majority of our depth information visually without any substantial effort or thought. The importance of depth information becomes more apparent when visual stimuli are diminished or absent. For example visually impaired individuals require a white cane or guide dog to scan their surroundings for obstacles. In the animal kingdom, there are numerous animal species that have evolved to live in low-light conditions or complete darkness and use echolocation to know where they are and what surrounds them. Without explicitly learning how, we are able to perceive the relative distances of objects in our environment.

Depth information is also important for technologies such as adaptive cruise control and autonomous emergency braking or for use in autonomous systems such as robots and self-driving vehicles. In these systems depth information can be used to decide whether to accelerate, brake or steer. Sonar, radar, and lidar\footnote{Radar that uses laser instead of radio waves} are examples of technologies that can be used to measure this information directly. As a complementary source of information or as a cost-effective alternative, depth can be predicted from camera data. Ground truth depth data collected using one of the aforementioned techniques can be used to train a depth estimation model in a supervised manner. As an alternative, similar to depth perception in the natural world, it is possible to train a depth estimation model using a self-supervised approach, using no ground truth depth information but only stereo image pairs \cite{garg_unsupervised_2016} or video data from a single camera \cite{zhou_unsupervised_2017}.


A single image depth prediction model can be trained using image reconstruction as the supervision signal. This reconstruction is done using image pairs, where the images are of the same scene, but taken from different positions. If a depth prediction is made for one of the images from such a pair, using information about the change of camera position, the first image can be reconstructed from the second image. By minimizing the difference between the original image and the reconstruction the task of depth prediction is learned.

The idea of learning depth by image reconstruction was used by \cite{garg_unsupervised_2016} on stereoscopic images with a known fixed camera transformation. \cite{zhou_unsupervised_2017} showed that it is possible to train depth prediction models on video data using image reconstruction as a supervision signal, by adding a parallel network that predicts the image-pair camera transformation that is required for image reconstruction. The reconstruction computation will be discussed in further detail in the methods section.

Compared to using stereoscopic images, learning this task from video has some unique challenges. Stereoscopic images can be taken simultaneously capturing a scene at a single point in time, whereas video frames are inherently captured at different points in time. If a video contains a dynamic scene (e.g. objects are moving) and this is not taken into consideration during training, moving objects will appear incorrectly in the reconstructed image and thereby  corrupt the learning signal.

The authors of \cite{zhou_unsupervised_2017} tried to tackle the dynamic scene problem by predicting a ‘motion explanation mask’ that can be used to ignore such regions. In later implementations of their work that are available online\footnote{\url{https://github.com/tinghuiz/SfMLearner}}, this mask was disabled, producing better results. In \cite{casser_depth_2018} , instance segmentation masks are used to handle object motion. These instance segmentation masks are only created for known object categories, using a pretrained mask  R-CNN model \cite{he_mask_2017}.

Another challenge is that the camera transformations between video frames vary, in contrast to the fixed camera-distance of stereo image pairs. When there is little to no camera movement, adjacent frames are nearly identical, and hardly provide any learnable information. \cite{zhou_unsupervised_2017} attempts to reduce this problem by filtering out the nearly identical frames from the training data.

When not taking object motion into account during training, a model may learn to make incorrect depth predictions to compensate for the reconstruction mistakes caused by object motion. For example, objects that during training were often observed while they were moving at the same velocity as the camera, will incorrectly be predicted as being far away.  This happens because the observed behavior of same-velocity-objects and distant stationary objects is similar; Their appearance does not change (much) from frame to frame. 
To counter this phenomenon \cite{godard_digging_2018}  applies a mask (called automask) to ignore pixels that do not change appearance from one frame to the adjacent frame. In addition of ignoring objects which appear static because they move with the same velocity as the camera, this mask will also ignore entire frames when the camera does not move.

\begin{figure}[htbp]
  \centering
  \resizebox{1.0\columnwidth}{!}{
  \newcommand{\turnheightnew}{0.195\columnwidth}
        \centering
        \begin{tabular}{@{\hskip 0mm}c@{\hskip 0mm}c}
            \includegraphics[width=0.8\columnwidth]{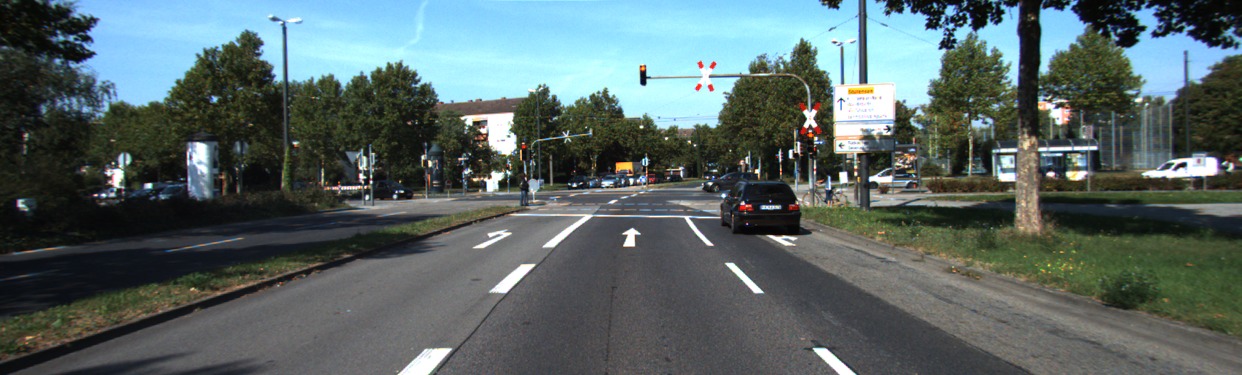} \\
            \includegraphics[width=0.8\columnwidth]{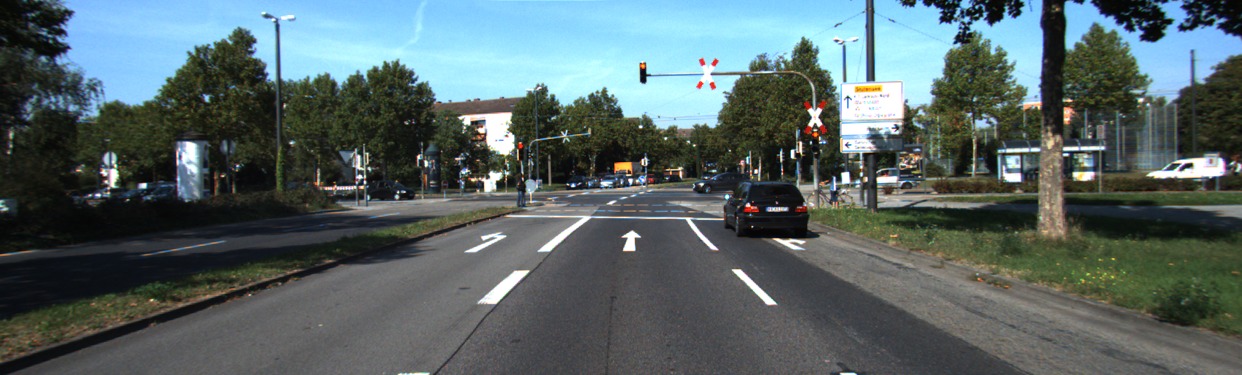} \\
            \includegraphics[width=0.8\columnwidth]{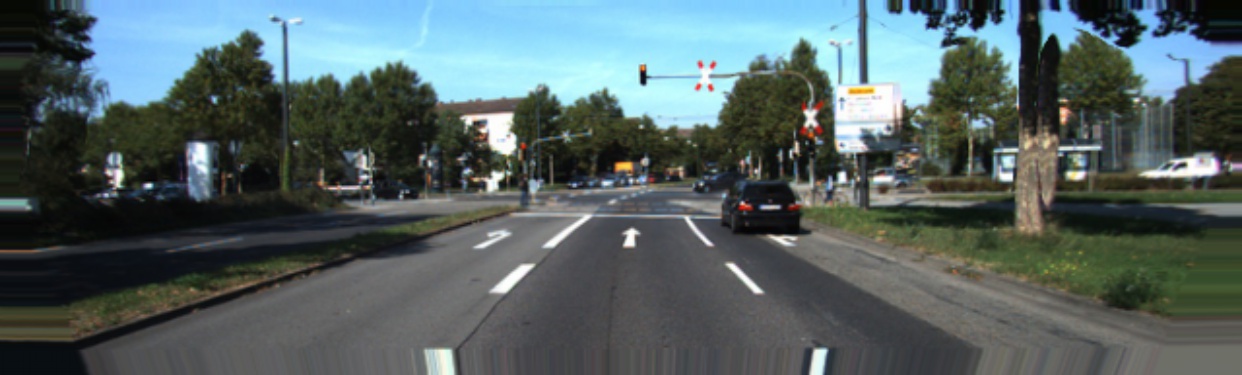}
        \end{tabular}
    }
  \caption{\textbf{Effects of occlusion on the reconstructed image.} \textbf{Top:} The image for which the depth prediction is made.
   \textbf{Middle:} The next video frame, captured 100ms later.
  \textbf{Bottom:} Reconstruction of the first image made by sampling from the next video frame. The lane markings at the bottom of the image are incorrectly reconstructed because they are not visible in the next frame. The tree on the right-hand side of the road appears twice in this reconstruction. This happens when areas with different predicted depths are projected to the same area. In other words, the sampling would have been correct if the tree was not blocking the view.
 }
    \vspace{-4pt}
  \label{fig:occlusion_example}
\end{figure}

One of the phenomena encountered when training a depth prediction model using image reconstruction is called occlusion. Occlusion relates to parts of a scene that can only be observed from one of two camera positions. For example, regions at the image boundaries may move in or out of view when the camera position changes. Another cause of occlusion is parallax, where for two different camera positions the apparent position of closer objects changes more, hiding or revealing  what is behind them. Both types of occlusion effects can be seen in figure \ref{fig:occlusion_example}. 
Incorporating these occlusion effects in the loss function during training can negatively impact the predictive performance of a depth prediction model since they do not provide meaningful information about the correctness of the depth predictions.

Various solutions have been proposed to ignore these occlusion effects during training. In \cite{mahjourian_unsupervised_2018} a mask is applied in the loss function to account for occlusion effects at the image boundaries. \cite{godard_digging_2018} tackles both types of occlusion at the same time using a loss function called \enquote{Per-pixel minimum reprojection loss}, which at each pixel location selects the best reconstruction and ignores the other ones. The idea behind this loss function is that when a pixel is occluded in the adjacent video frame, a correct reconstruction of the pixel is unlikely, and the reconstruction of that pixel will not be used to optimize the depth prediction model.

\begin{figure}[htbp]
  \centering
    \begin{subfigure}{0.49\columnwidth}
        \includegraphics[width=\textwidth]{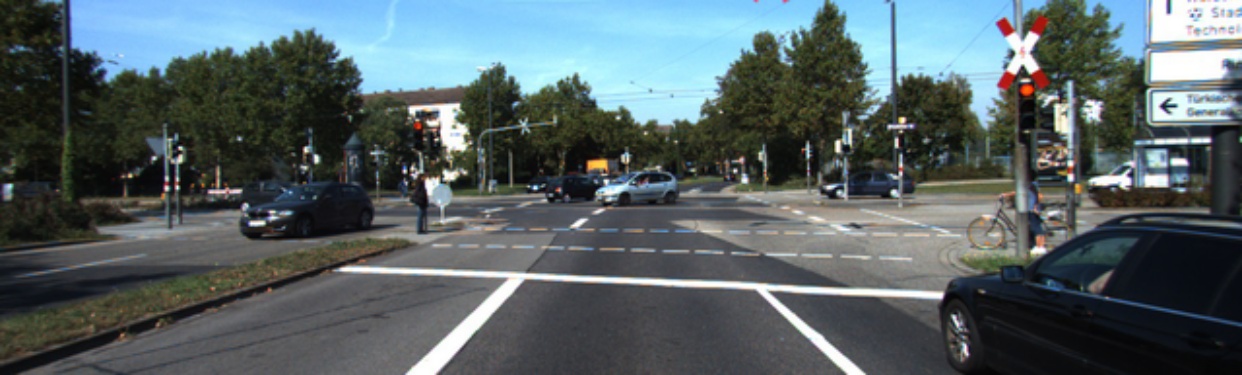}
    \end{subfigure}

    \begin{subfigure}{0.49\columnwidth}
    \includegraphics[width=\textwidth]{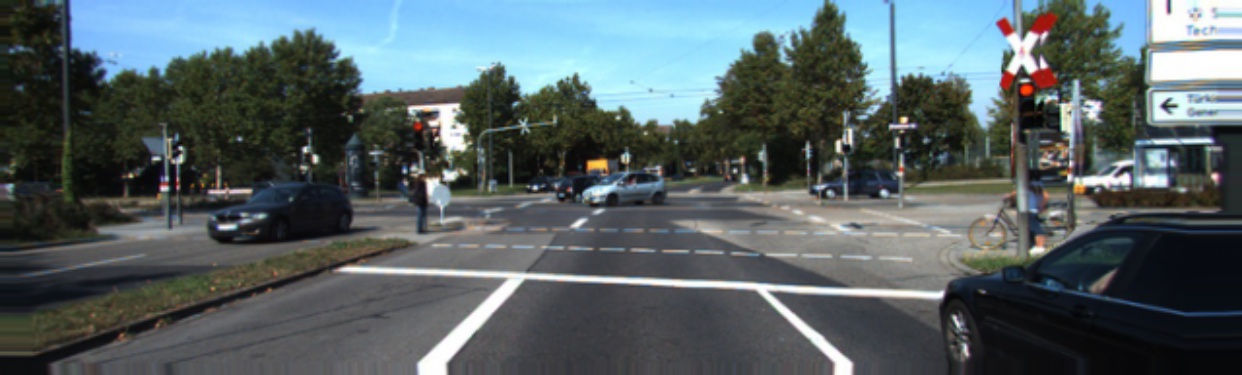}
    \end{subfigure}
    \begin{subfigure}{0.49\columnwidth}
    \includegraphics[width=\textwidth]{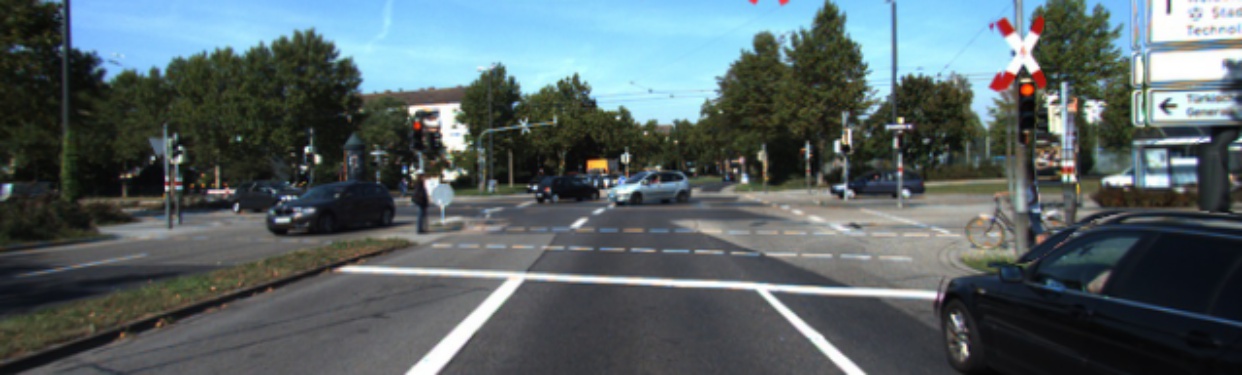}
    \end{subfigure}
    
    \begin{subfigure}{0.49\columnwidth}
    \includegraphics[width=\textwidth]{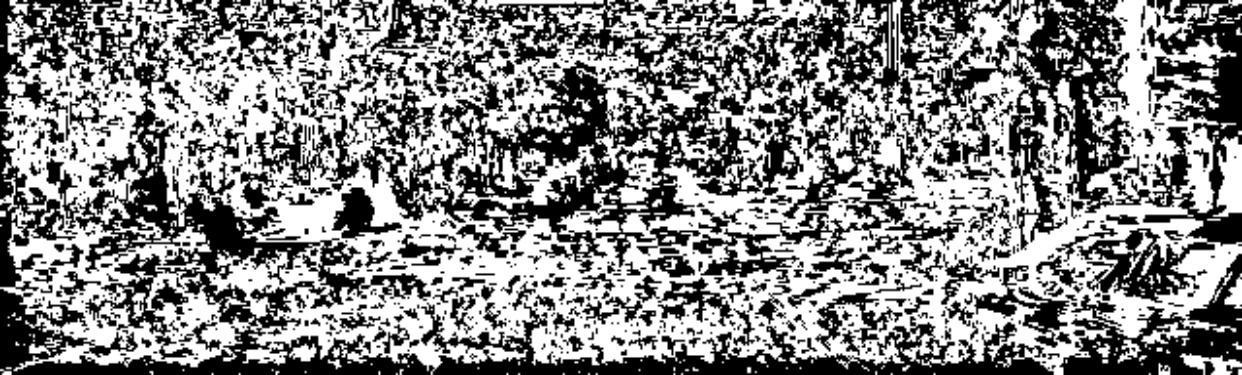}
    \end{subfigure}
    \begin{subfigure}{0.49\columnwidth}
    \includegraphics[width=\textwidth]{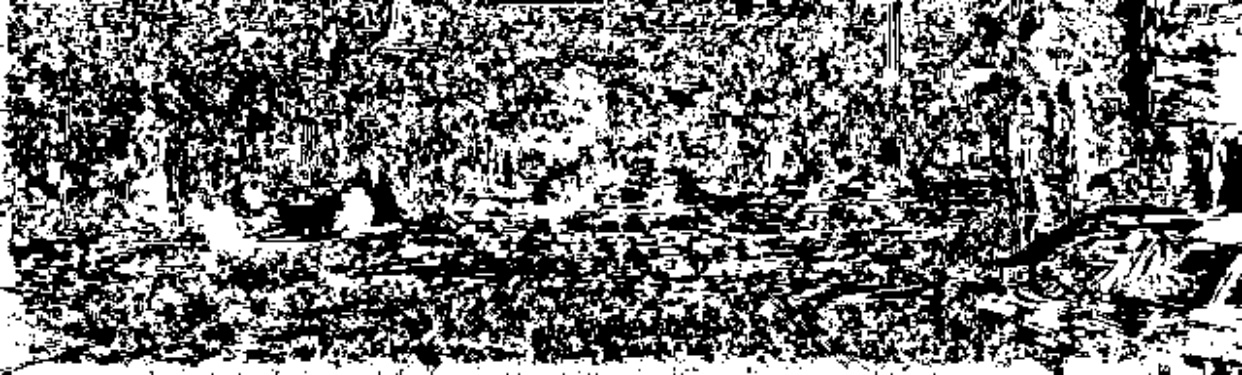} 
    \end{subfigure}
    
    \begin{subfigure}{0.49\columnwidth}
    \includegraphics[width=\textwidth]{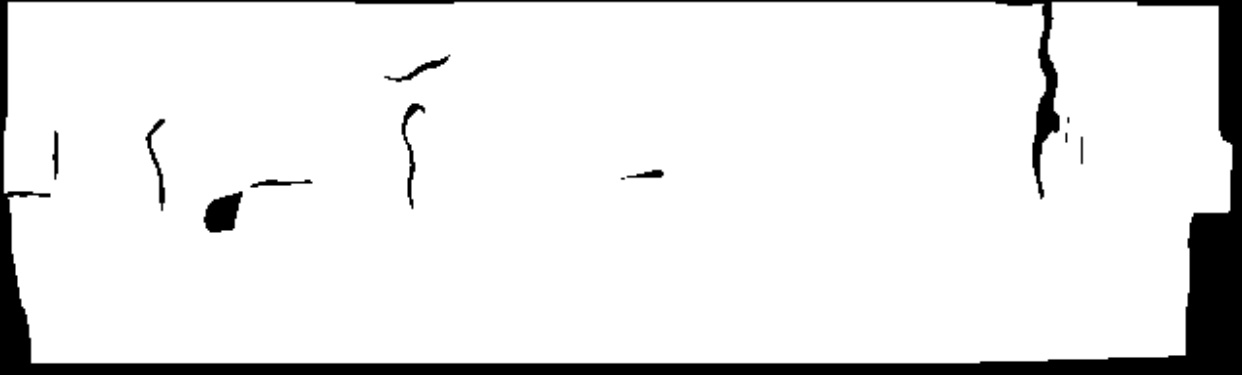}
    \end{subfigure}
    \begin{subfigure}{0.49\columnwidth}
    \includegraphics[width=\textwidth]{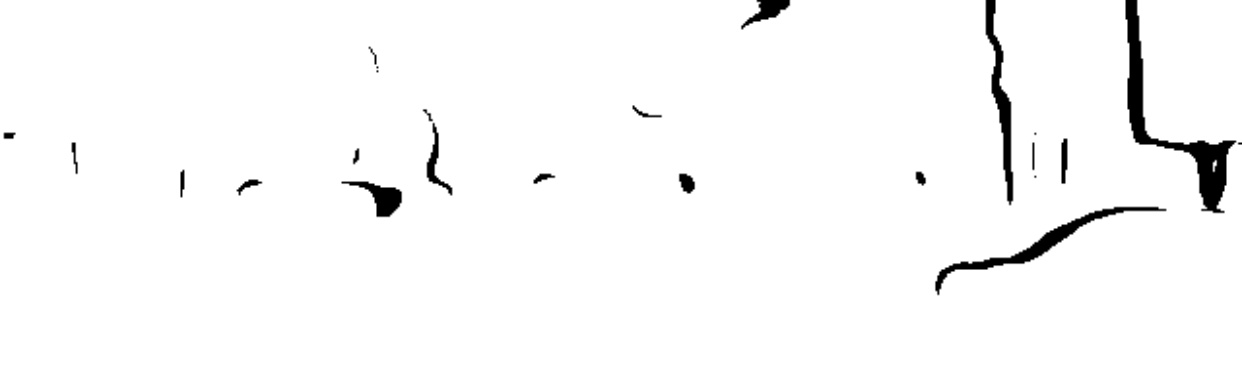} 
    \end{subfigure}
            
  \caption{
\textbf{Top:} Image for which the depth prediction is made and target for image reconstruction. 
\textbf{Second row:} Reconstructions of the above image, made from the next and previous video frames respectively. 
\textbf{Third row:} Map of each reconstructions pixels that are used/ignored (white/black) by minimum reprojection loss. 
\textbf{Bottom row:} Examples of occlusion mask for each reconstruction, black areas are affected by occlusion and should not be used in the loss function.
 }
    \vspace{-4pt}
  \label{fig:ppmrl_example}
\end{figure}

Since the per-pixel minimum reprojection loss selects which reconstruction will be used per pixel location, it effectively creates a binary mask for each of the reconstructions. This is illustrated in figure \ref{fig:ppmrl_example}. The frame which is the target for image reconstruction is shown at the top of the figure. On the second row of the figure, two reconstructed images are shown, made from the next and previous video frames respectively. A binary map\footnote{Using the word \enquote{map}, instead of \enquote{mask} since it is not actively used to mask.}is located below each of these reconstructions. These maps display for each pixel of the reconstruction above it whether it is used or ignored by the per-pixel minimum reprojection loss.

From these binary maps, it can be observed that regions which are mostly black coincide with occlusion effects.  This indicates that per-pixel minimum reprojection loss is able to ignore occlusion effects successfully. What also can be observed is that the pixels of the reconstruction target image for which the contents are visible in both adjacent frames, still only one of the reconstructions is selected to be used in the loss, and the other one is ignored. 

The hypothesis that I will test in this work is that by limiting the amount of information which is ignored during training to only those regions that suffer from occlusion, more useful information will be available for training the model. 
The approach I propose is \textit{occlusion mask}, a mask that during training can be used to specifically discard regions that cannot be reconstructed due to occlusions. The images at the bottom row of figure \ref{fig:ppmrl_example} illustrate the concept. Occlusion mask is based entirely on depth information.
We will test the hypothesis by incorporating the proposed occlusion mask method into two novel loss functions, use these loss functions to train depth prediction models, and compare their  performance with that of a model trained with per-pixel minimum reprojection loss. 
Additionally we will compare the behavioral differences of the loss functions, i.e. those that incorporate the proposed method and the per-pixel minimum reprojection loss, by visualizing the computed loss on training examples. With the occlusion mask we hope to improve the performance of future single image depth prediction models.


\section{Method}
We start by reviewing the framework introduced by \cite{zhou_unsupervised_2017} for training a single-view depth network from unlabeled video data, including its core concept: image reconstruction.
This will be followed by a review of the loss function computation of the method and implementation\footnote{\url{https://github.com/nianticlabs/monodepth2}} by Godard et al \cite{godard_digging_2018}, which serves as the foundation for our modifications as well as the baseline in our experiments.
We will finish the method section by describing our proposed occlusion mask and introducing two novel loss formulations which incorporate it. 

\subsection{Self-supervised training}
\subsubsection{Framework}

\begin{figure}[htbp]
  \centering
  \resizebox{1.0\columnwidth}{!}{
        \centering
            \includegraphics[width=0.8\columnwidth]{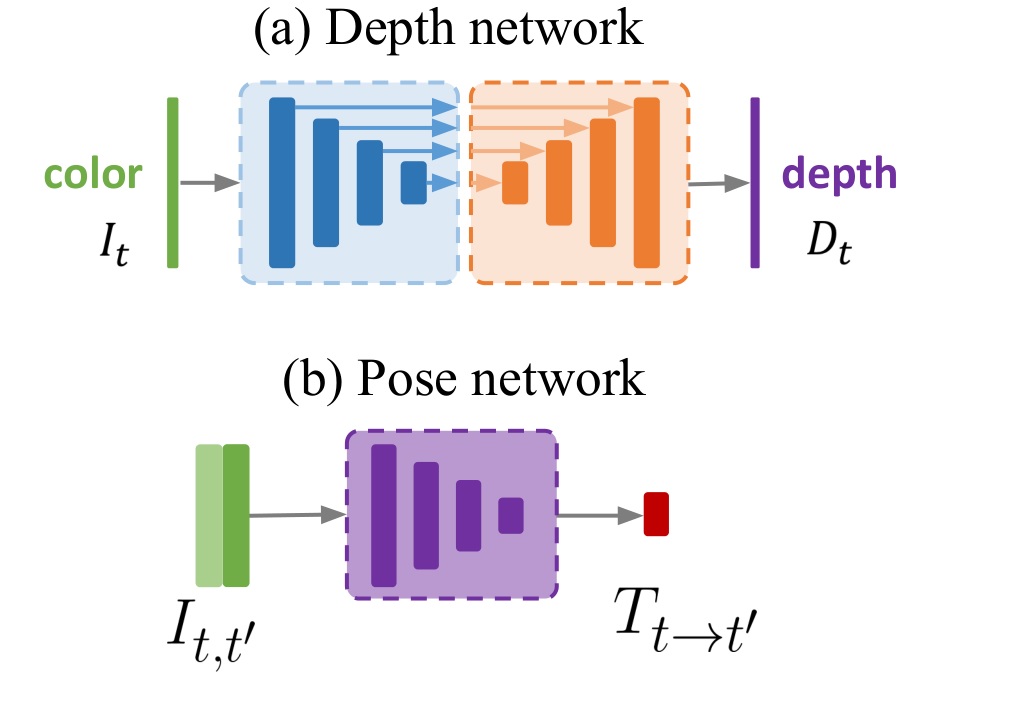}
           
    }
  \caption{\textbf{(a) Depth network} A standard, fully convolutional, U-Net is used to predict depth. \textbf{(b) Pose network} Pose between a pair of frames is predicted with a separate pose network.  Image from \cite{godard_digging_2018}.
 }
  \label{fig:network}
\end{figure}

All of the models in our experiments are trained using the same framework  introduced by \cite{zhou_unsupervised_2017} consisting of one network predicting a depth map from a single image (i.e. in this case a frame from a video), and a pose network predicting camera transformation from two consecutive images (figure \ref{fig:network}). The two networks are trained together using the same loss but can be used separately after training. 

The loss used for training is based on warping nearby frames, to create a new image that conforms (as best as possible) to the image for which the depth map is being predicted.
This ‘warping’ means that each pixel is sampled at their new location in the nearby frame, which can be computed using the predictions of the depth map and camera transformation.

\subsubsection{Image reconstruction}
\label{image_reconstruction}

A frame is reconstructed by sampling from one of the adjacent video frames. 
Equation \ref{eq:sampling} shows how the sample location in the adjacent frame is computed for each pixel in the reconstruction. In this equation the subscript is used to indicate the point in time, $t$ is “now”, i.e. the time of the frame for which when the depth is predicted, $t'$ is the time point of the next or previous frame, and $t \to t'$ is the transition from one to the other.

The used formulation of the image reconstruction has the following implicit assumptions \cite{zhou_unsupervised_2017}: 
1) the scene is static without moving objects; 
2) there is no occlusion/disocclusion between the target view and the source views; 
3) the surfaces appear uniformly bright from all directions of view so that the photo-consistency error is meaningful

\begin{equation}
\label{eq:sampling}
\begin{pmatrix} x_{t \to t'}z_{t \to t'} \\ y_{t \to t'}z_{t \to t'} \\ z_{t \to t'} \\ 1 \end{pmatrix} = K T_{t \to t'} K^{-1} \begin{pmatrix} x_{t }z_{t} \\ y_{t}z_{t} \\ z_{t} \\ 1 \end{pmatrix}
\end{equation}

We will first describe equation \ref{eq:sampling} and follow with an example that illustrates the computational steps. The purpose of this equation is to compute the sampling location $(x_{t \to t'}, y_{t\to t'})$ in frame $I_{t'}$ for each pixel at position $(x_t, y_t)$ in frame $I_t$. In this equation the rightmost column vector consists of the homogeneous coordinates of a pixel in frame $I_t$ where $z_t$ is the predicted depth of that pixel. The leftmost column vector consists of the homogeneous coordinates of the pixel after projection, i.e. the location at which the pixel is to be expected in frame $I_{t'}$ given the predicted depth and camera transformation. The 4x4 camera transformation matrix $T_{t\to t'}$ contains the rotation and translation of the camera as predicted by the pose network. The matrices $K$ and $K^{-1}$ denote the camera intrinsics matrix and its inverse, which transforms camera coordinates to image coordinates using camera properties such as focal length and principal point offset.\footnote{ To get an idea of how the camera transformation matrix and the camera intrinsics matrix affect what is seen in an image, you can have a look at this interactive tool: \url{https://ksimek.github.io/perspective_camera_toy.html}}

\begin{figure}[htbp]
    \centering
    \begin{subfigure}{0.49\columnwidth}
        \includegraphics[width=\textwidth]{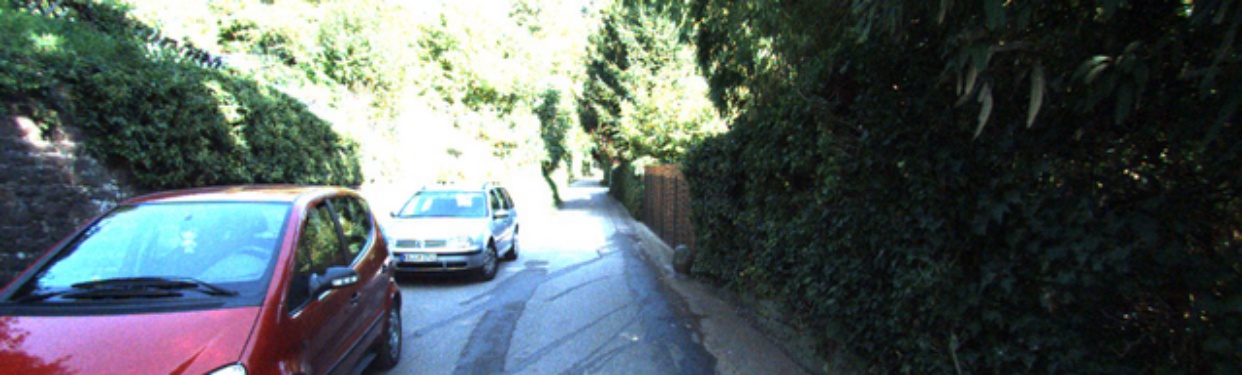}
        \subcaption{Target image} 
        \label{fig:reproject_original}
    \end{subfigure}
    \begin{subfigure}{0.49\columnwidth}
        \includegraphics[width=\textwidth]{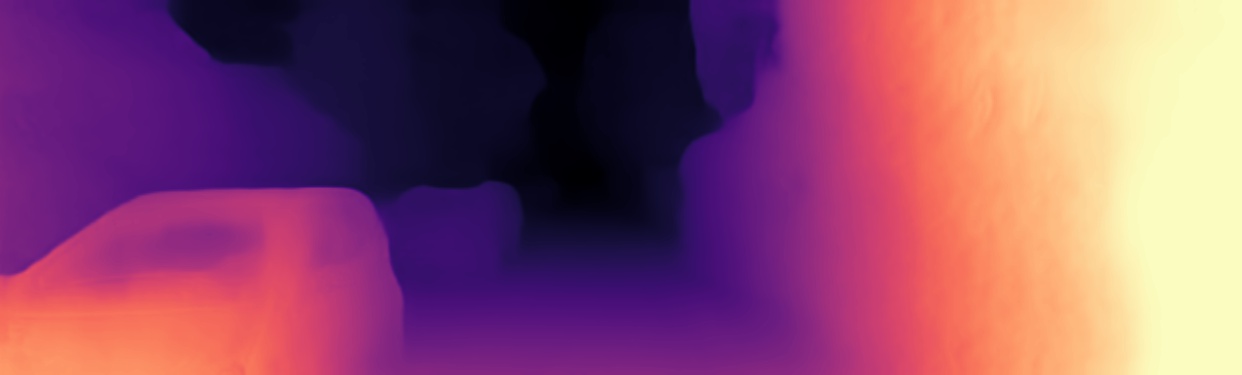}
        \caption{Predicted depth map} 
        \label{fig:reproject_disp}
    \end{subfigure}
    
    \begin{subfigure}{0.49\columnwidth}
        \includegraphics[width=\textwidth]{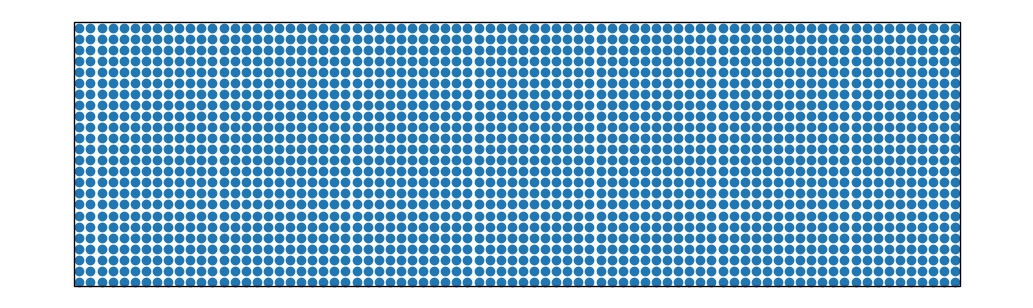} 
        \caption{Grid of pixel locations} 
        \label{fig:reproject_grid}
    \end{subfigure}
    \begin{subfigure}{0.49\columnwidth}
        \includegraphics[width=\textwidth]{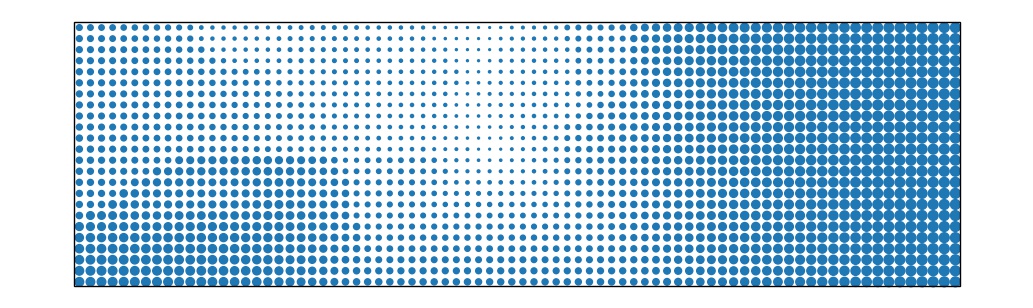} 
        \caption{Point cloud = grid + depth} 
        \label{fig:reproject_grid_depth}
    \end{subfigure}
    
    \begin{subfigure}[t]{0.49\columnwidth}
        \includegraphics[width=\textwidth]{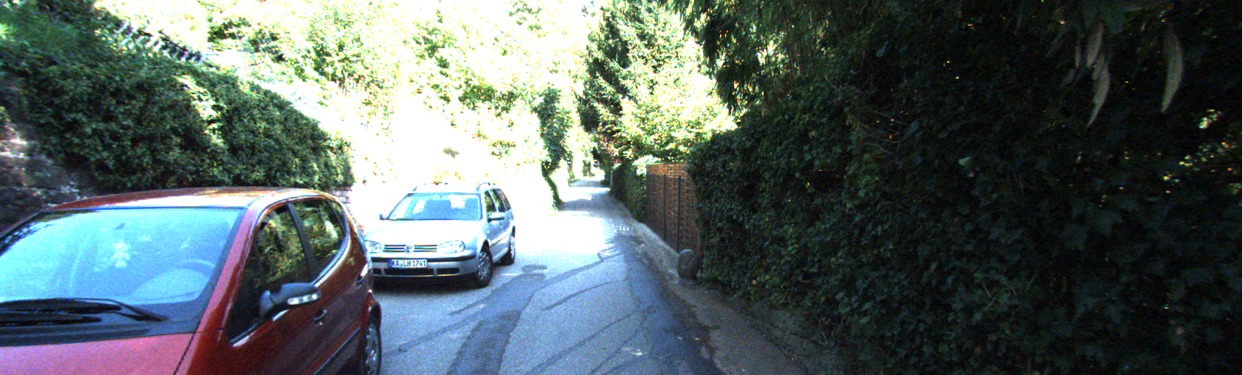} 
        \caption{Next video frame} 
        \label{fig:reproject_next_frame}
    \end{subfigure}
    \begin{subfigure}[t]{0.49\columnwidth}
        \includegraphics[width=\textwidth]{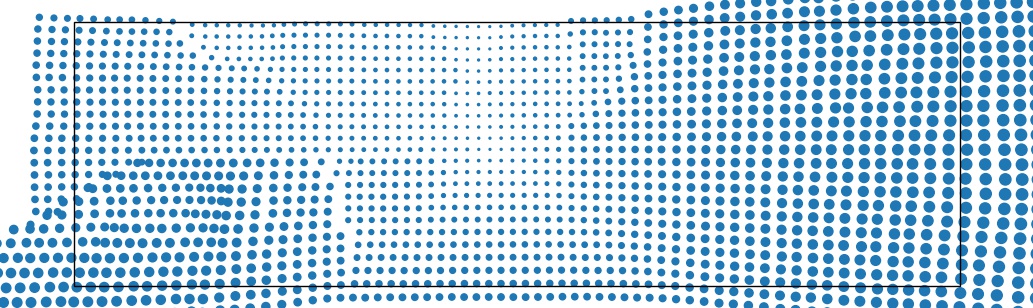} 
        \caption{Point cloud projected to next frame} 
        \label{fig:reproject_grid_proj_to_next}
    \end{subfigure}
    
    \begin{subfigure}{0.49\columnwidth}
        \includegraphics[width=\textwidth]{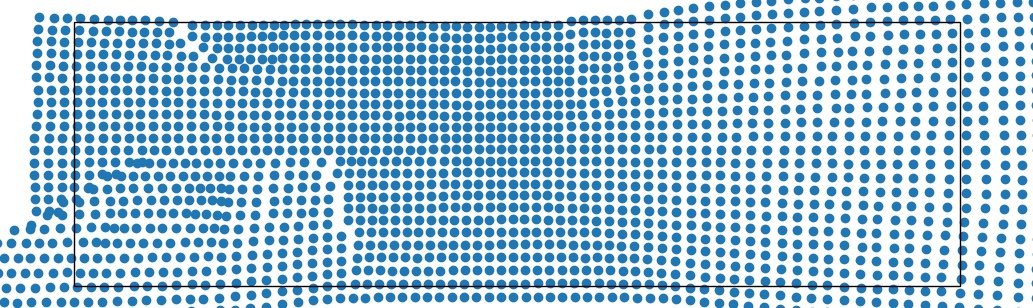}
        \caption{Sample locations} 
        \label{fig:reproject_grid_proj_to_next_no_depth}
    \end{subfigure}
    \begin{subfigure}{0.49\columnwidth}
        \includegraphics[width=\textwidth]{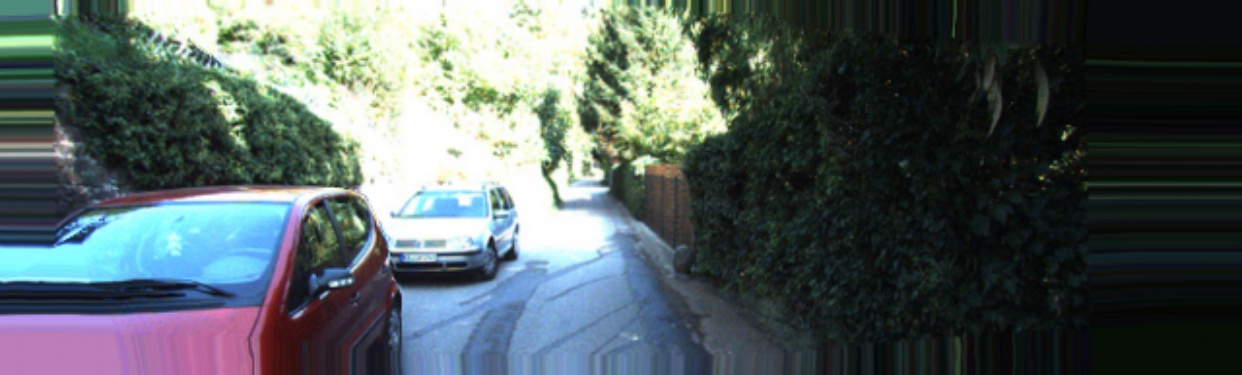}
        \caption{Reconstructed image} 
        \label{fig:reproject_proj_from_next}
    \end{subfigure}
            
    \caption{
    \textbf{Example of image reconstruction steps.}
    For an image (a), a depth map prediction (b\&d) together with a camera transformation prediction is used to compute the sample locations (g) in the adjacent image (e) that can be used to reconstruct (h) the target image .
 }
    \vspace{-4pt}
  \label{fig:reproject_example}
\end{figure}

In figure \ref{fig:reproject_original} \& \ref{fig:reproject_disp} you can see a picture and its predicted depth map. Figure \ref{fig:reproject_grid} shows the outline of an image (the black rectangle) and its pixels neatly arranged in a grid (the blue dots). The depth information of \ref{fig:reproject_disp} can be added to each pixel in \ref{fig:reproject_grid} to create the point cloud shown in \ref{fig:reproject_grid_depth}, where the dots are bigger when the point is closer to the camera. \ref{fig:reproject_next_frame} shows the frame that comes next after frame \ref{fig:reproject_original}. The camera transformation (from \ref{fig:reproject_original} to \ref{fig:reproject_next_frame}) predicted by the pose network can be used to calculate where the points of \ref{fig:reproject_grid_depth} will end up, this is shown in \ref{fig:reproject_grid_proj_to_next}. Figure \ref{fig:reproject_grid_proj_to_next_no_depth} shows the points from \ref{fig:reproject_grid_proj_to_next} but with the depth information removed. Frame \ref{fig:reproject_original} can be reconstructed by sampling at the new pixel positions (\ref{fig:reproject_grid_proj_to_next_no_depth}) in \ref{fig:reproject_next_frame}. Figure \ref{fig:reproject_proj_from_next} shows this reconstruction. 

Effects of the sample locations can be seen in the reconstruction, for example pixels where the sample location is outside of the image (the black rectangle) the closest border pixel is used instead. Another example is the small area near the tail light of the (red) car from which there is no sampling as can be seen in \ref{fig:reproject_grid_proj_to_next_no_depth}. This area in \ref{fig:reproject_next_frame} shows the headlamp of the car in the rear, which is not visible (occluded) in the target image \ref{fig:reproject_original}.

The sample coordinates $(x_{t \to t'}, y_{t \to t'})$ are continuous values. The reconstructed frame $I_{t' \to t}$ can be made by sampling at these projected pixel coordinates in frame $I_{t'}$ using the (sub-)differentiable bilinear sampling mechanism proposed in \cite{jaderberg_spatial_2015}. This sampling method linearly interpolates the values of the four pixels that surround the sample location.

\subsection{Baseline loss function} 
In this section we will discuss the components of the loss function used in the baseline \cite{godard_digging_2018}. The loss used for training (eq. \ref{eq:loss}) consists of two components, which will be discussed in this section. One component is the smoothness loss ($L_s$) over the predicted depth map. This loss remains unchanged in the experiments, and is scaled with smoothness term $\lambda$ set to 0.001. 
The other component of the loss function is the photometric loss multiplied with a binary mask in order to ignore certain areas of a reconstruction. This mask ($\mu$ in the equation), called automask is the same in all of the experiments. The photometric loss ($L_p$) is computed differently in each of the experiments. 

\begin{equation}
\label{eq:loss}
L = \lambda L_s + \mu L_p
\end{equation}

The final loss is averaged over each pixel location, the various scales, and the images in a batch.

\subsubsection{Multiple scales} 
Due to the sampling mechanism described in \ref{image_reconstruction}, the gradient is derived from the difference between the target pixel value and the pixel value in the reconstruction which is an interpolation of the four pixels surrounding the sample location. Using this gradient will mean that the depth prediction is changed in the direction that moves the sample location closer to one of the four pixels that surround the current sample location and that matches the target pixel most closely.

This change in depth prediction is not necessarily in the right direction, for example when there is no gradual color transition of the pixels that lie between the current sample location and the correct sample location. This can happen if the current sample location is far from the correct sample location, or when the scene is complex e.g. many surfaces with different color gradients, or multiple color gradients in a single surface.

To prevent getting stuck in these local optima, the loss (and therefore also the gradient) is usually computed on multiple scales which allows the gradient to be derived from larger spatial regions directly.\cite{zhou_unsupervised_2017} This is done by making reprojections using the intermediate depth map predictions of the network, which have a lower resolution.

The authors of  \cite{godard_digging_2018} observed that projecting the input images with the resolution of the depth map, has the tendency to create artefacts when the depth map resolution is low and there are large low-texture regions in the image. They overcome this problem by upsampling the predicted depth maps to the resolution of the input image instead of downsampling the input image to the resolution of the depth map.\cite{godard_digging_2018} This preserves the details in low-texture regions which reduces ambiguity. Using this method values in the lower resolution depth map will influence a larger spatial region in the higher resolution reconstruction. 

\subsubsection{Smoothness loss}
There are many incorrect depth map predictions which could also provide accurate image reconstructions, for example in homogeneous regions of an image. To steer the network to learn more realistic predictions a loss is used that enforces smoothness of the predicted depth map. In this implementation an edge-aware smoothness loss is used (eq. \ref{eq:smoothness_loss}) \cite{godard_unsupervised_2016}. 
\begin{equation}
\label{eq:smoothness_loss}
L_s =  | \partial_x d^*_t | e^{- | \partial_x I_t |} + |\partial_y d^*_t | e^{-| \partial_y I_t |} 
\end{equation}
In this formulation depth discontinuities, i.e. high disparity gradients ($| \partial d^*_t |$), contribute less to the loss when the image gradients ($| \partial I_t |$) are relatively high. The used (inverse) depth map is normalized ($d_t^*$) by its mean value. This normalization is done to prevent that the loss is minimized by lowering the depth gradients through scaling down the entire depth map. This would be possible since the other loss component (i.e. the photometric error) is unaffected by scale, the depth and pose network would together scale down their predictions which negatively affects training. \cite{wang_learning_2018}

\subsubsection{Automask}
A mask, called automask, is used to ignore the stationary pixels, which come either from objects that move with the same velocity as the observer or when the camera is not moving. It is supposed to prevent the pixels which remain stationary in the image from contaminating the loss. It does this by ignoring the loss of pixels where the photometric error of the original, unwarped frame $I_t'$ is lower than the warped frame $I_{t' \to t}$. Equation \ref{eq:automask} shows the mask definition where $pe$ stands for photometric error, which will be described in the next section.
\begin{equation}
\label{eq:automask}
\mu  = \begin{cases} 1 & \text{if } \displaystyle \min_{t'} pe(I_t, I_{t'}) > \displaystyle \min_{t'} pe(I_t, I_{t' \to t}) \\ 0 & \text{otherwise} \end{cases}
\end{equation}
\subsubsection{Photometric error}
\begin{figure}[htbp]
    \centering
    \begin{subfigure}{\columnwidth}
        \includegraphics[width=\textwidth]{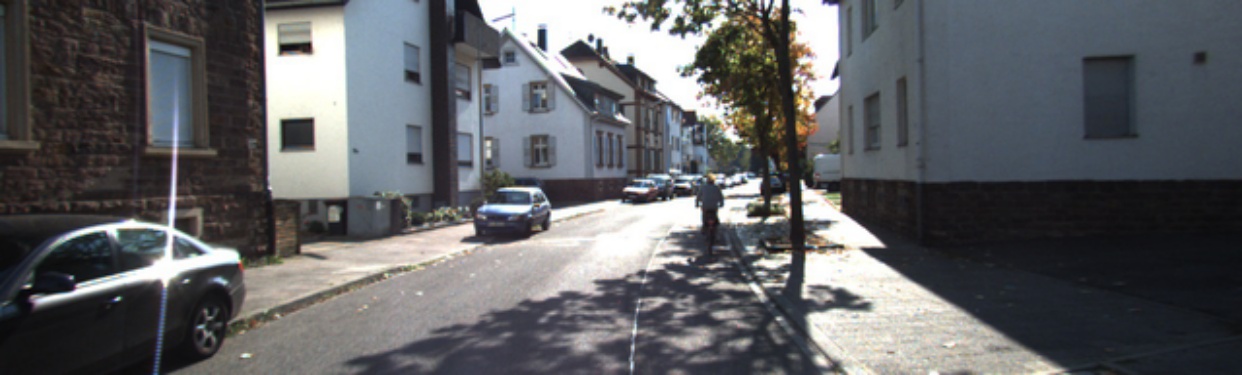}
    \end{subfigure}
    \begin{subfigure}{\columnwidth}
        \includegraphics[width=\textwidth]{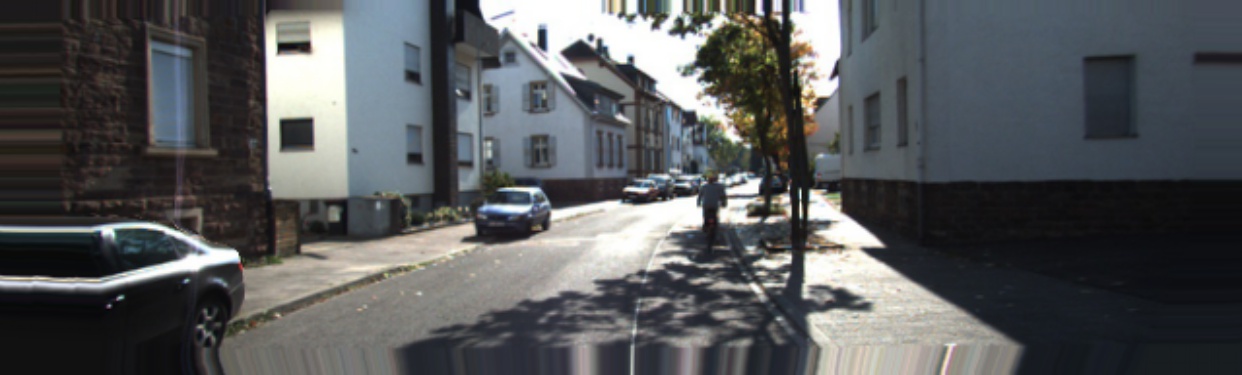}
    \end{subfigure}
    
    \begin{subfigure}{\columnwidth}
        \includegraphics[width=\textwidth]{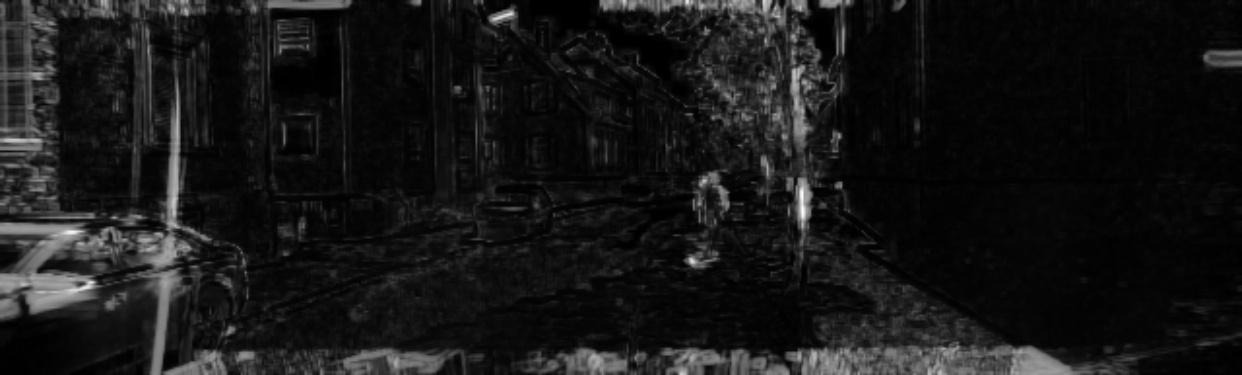}
    \end{subfigure}
    
    \caption{\textbf{Example of the photometric error} \textbf{Top:} The reconstruction target image. \textbf{Middle:} Reconstruction made from the next video frame. \textbf{Bottom:} The calculated photometric error between the target and the reconstruction.
}
  \label{fig:photometric_error_example}
\end{figure}

The photometric error (eq. \ref{eq:photometric_error}) used in the photometric loss component is a combination of the L1 loss of the pixel value differences and the structural similarity (SSIM) index \cite{zhou_wang_image_2004} of the image for which the depth is predicted and its reconstruction made from the adjacent frame . SSIM is used because it is a measure of structural information change and the human visual system is adapted to extract structural information. In the experiments $\alpha = 0.85$ is used. Figure \ref{fig:photometric_error_example} shows an example of the photometric error (\ref{fig:photometric_error_example}c) computed on the target “reference” image (\ref{fig:photometric_error_example}a) and the reprojected image (\ref{fig:photometric_error_example}b).
\begin{equation}
\label{eq:photometric_error}
pe(I_a, I_b) = \alpha \frac{1 - SSIM(I_a, I_b)}{2} + (1 - \alpha) \| I_a - I_b\|_1 
\end{equation}

\subsubsection{Per-Pixel Minimum Reprojection Loss}
\label{ppmr_loss}
In the baseline per-pixel minimum reprojection loss (equation \ref{eq:ppmr_loss}) is used as the photometric loss . Which means that at each pixel location, the reconstruction with the lowest error is used in the loss. This is different to previous self-supervised depth estimation methods which instead use the average of the reconstructions.

Selecting the reconstruction with lowest error at that pixel location is done to account for regions in the reconstruction target (image), that are not visible in some of the images that are used to make the reconstructions. Even when the depth of these regions is predicted correctly, a correct reconstruction is not likely, which would give a high photometric error for such a region. \cite{godard_digging_2018} 
\begin{equation}
\label{eq:ppmr_loss}
L_p = \min_{t'} pe(I_t, I_{t'\to t})
\end{equation}
\subsection{Occlusion}
\subsubsection{The occlusion mask}

The hypothesis in this thesis is that by selectively ignoring only the occluded regions of a reconstructed image, compared with the per-pixel minimum reprojection loss, more relevant information is available to learn the task of depth prediction. To this end we design an occlusion mask for discarding regions that cannot be reconstructed due to occlusions. Our solution for the occlusion mask utilizes the previously unused depth information of the point/pixel cloud after it has been projected in order to reconstruct the target image (figure \ref{fig:reproject_grid_proj_to_next} in section \ref{image_reconstruction}). 

\begin{figure*}[htbp]
    \centering
    \begin{subfigure}{0.30\linewidth}
        \includegraphics[width=\textwidth]{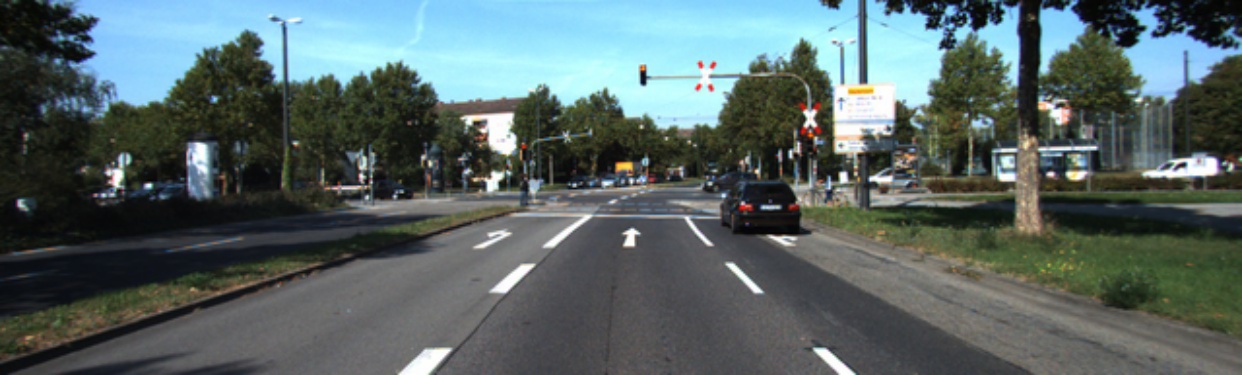}
        \caption{Target image} 
        \label{fig:occlusion_mask_original}
    \end{subfigure}
    \begin{subfigure}{0.30\linewidth}
        \includegraphics[width=\textwidth]{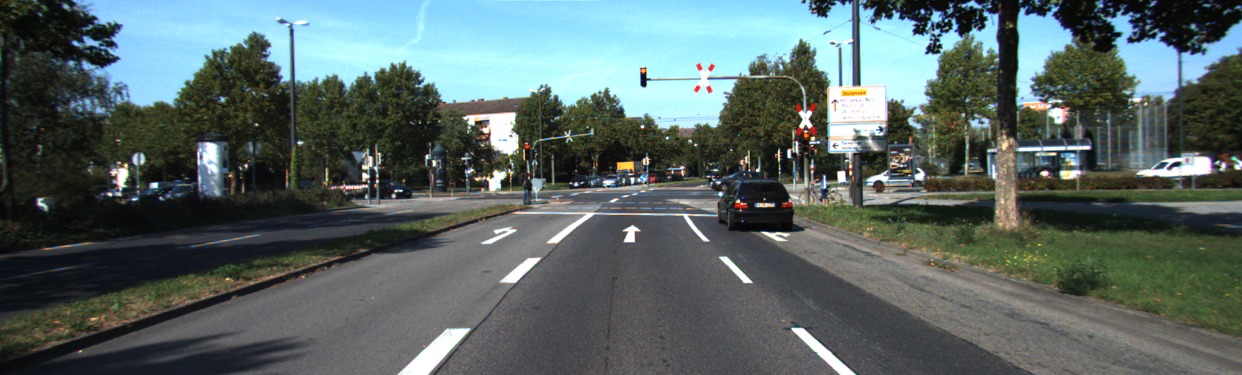}
        \caption{Previous video frame} 
        \label{fig:occlusion_mask_previous_frame}
    \end{subfigure}
    \begin{subfigure}{0.30\linewidth}
        \includegraphics[width=\textwidth]{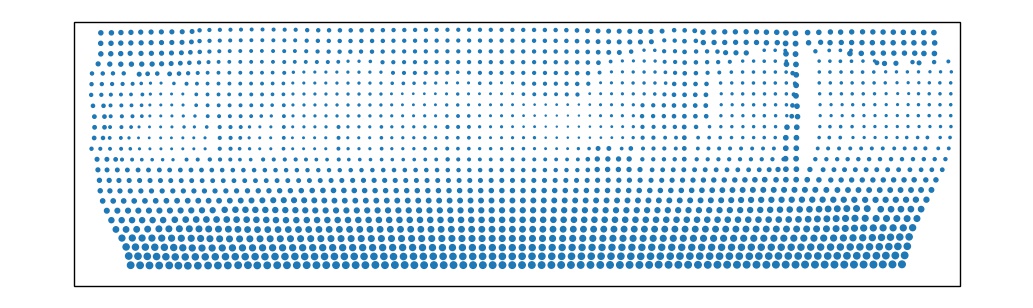} 
        \caption{Projected point cloud} 
        \label{fig:occlusion_mask_grid_project_to_previous}
    \end{subfigure}
    \begin{subfigure}{0.30\linewidth}
        \includegraphics[width=\textwidth]{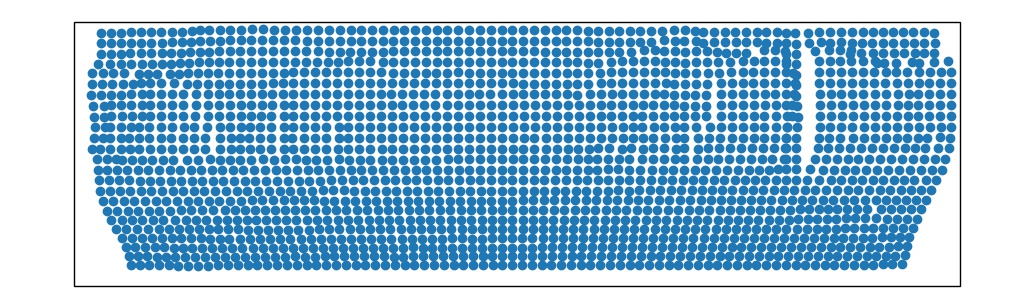}
        \caption{Sample locations} 
        \label{fig:occlusion_mask_grid_project_to_previous_no_depth}
    \end{subfigure}
    \begin{subfigure}[t]{0.30\linewidth}
        \includegraphics[width=\textwidth]{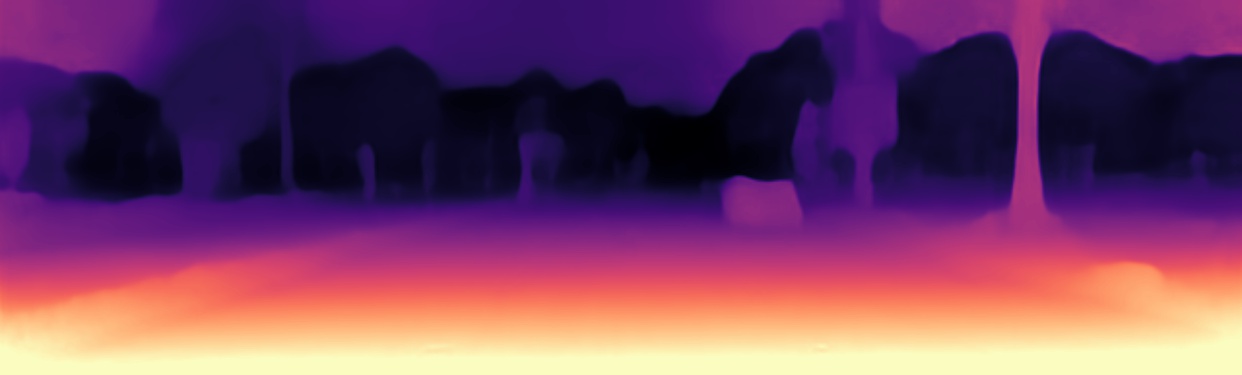}
        \caption{Depths from projected point cloud} 
        \label{fig:occlusion_mask_expected_sample_depth}
    \end{subfigure}
    \begin{subfigure}[t]{0.30\linewidth}
        \includegraphics[width=\textwidth]{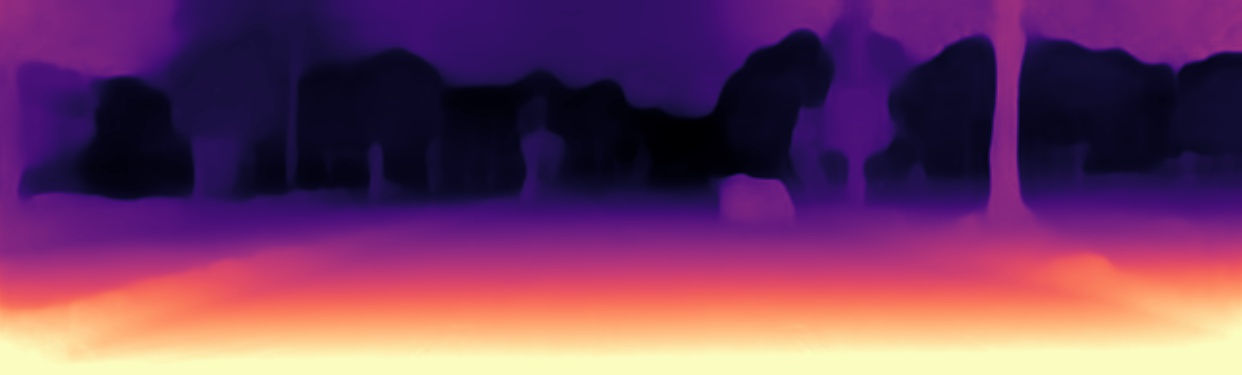}
        \caption{Predicted depth map of previous frame} 
        \label{fig:occlusion_mask_disp_previous_frame}
    \end{subfigure}
    \begin{subfigure}{0.30\linewidth}
        \includegraphics[width=\textwidth]{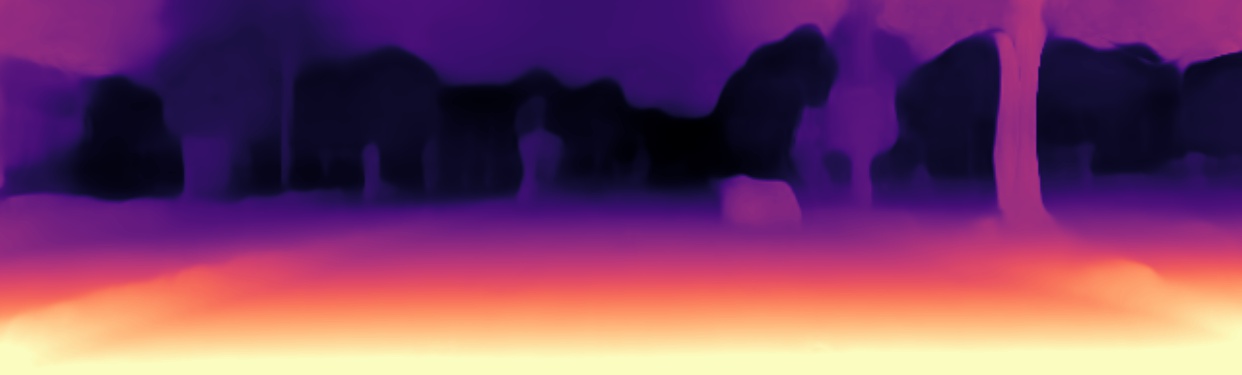}
        \caption{Depths sampled from \ref{fig:occlusion_mask_disp_previous_frame}} 
        \label{fig:occlusion_mask_sampled_disp_previous_frame}
    \end{subfigure}
    \begin{subfigure}{0.30\linewidth}
        \includegraphics[width=\textwidth]{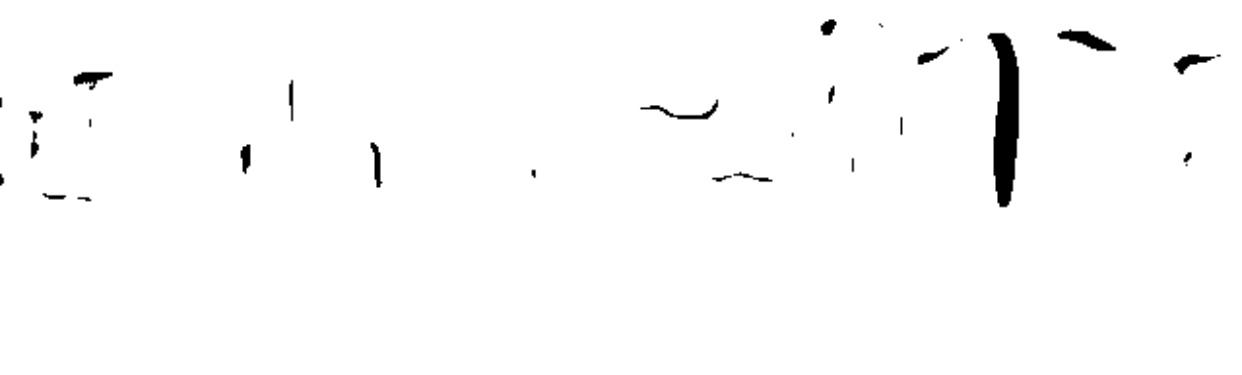}
        \caption{Occlusion mask} 
        \label{fig:occlusion_mask_mask}
    \end{subfigure}
    \begin{subfigure}{0.30\linewidth}
        \includegraphics[width=\textwidth]{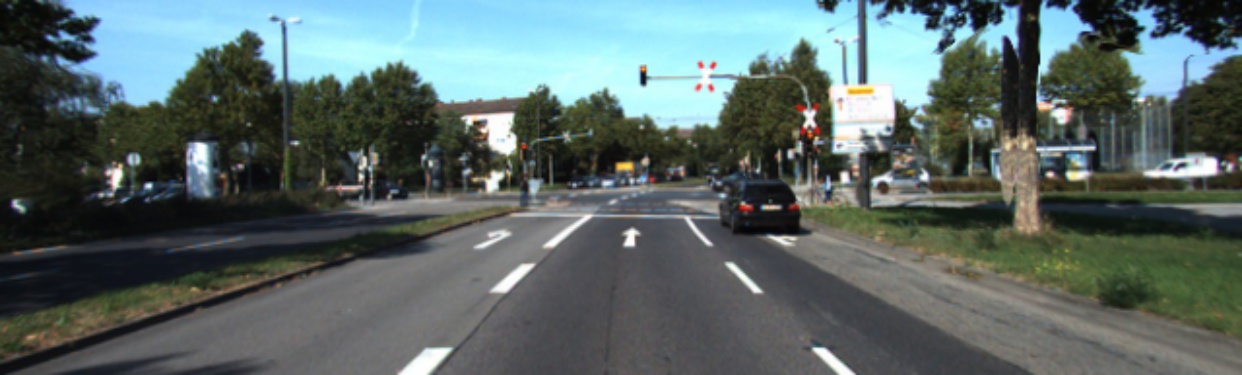}
        \caption{Reconstructed image} 
        \label{fig:occlusion_mask_proj_from_previous}
    \end{subfigure}
            
    \caption{\textbf{Example of occlusion mask computation steps.} 
    To make a reconstruction (i) of frame (a) from an adjacent frame (b) , sample locations (d) are computed. This calculation also provides each sample locations expected depth (e). These expected depths (e) can be compared with the depths observed at the sample locations (g), i.e. the depths sampled from the adjacent frame depth map prediction(f). This comparison gives the occlusion mask (h), which shows an occlusion when the observed depth at the sample location is closer than the depth that is expected.
 }
    \vspace{-4pt}
  \label{fig:occlusion_mask_example}
\end{figure*}

Figure \ref{fig:occlusion_mask_example}  illustrates the computation steps of the occlusion mask. Consider  \ref{fig:occlusion_mask_original} as the frame for which the depth map is predicted and the target of the reconstruction, and \ref{fig:occlusion_mask_previous_frame} the previous video frame. Similar to \ref{fig:reproject_grid_proj_to_next} from the example in section \ref{image_reconstruction} , figure \ref{fig:occlusion_mask_grid_project_to_previous} shows the projected points, i.e. where the pixels of the reconstruction target image will end up after the camera transformation to the previous video frame. In contrast with that earlier example, we will use the depth information that is removed from the point cloud visible in \ref{fig:occlusion_mask_grid_project_to_previous} to get the sample locations \ref{fig:occlusion_mask_grid_project_to_previous_no_depth}. This information gives the depths that are expected at the sample locations. 

These depth expectations are visualized in \ref{fig:occlusion_mask_expected_sample_depth}, important to note is that it is not an actual depth map of a point in time, but the depths that you would expect to see if you would sample at the locations of \ref{fig:occlusion_mask_grid_project_to_previous_no_depth} in the depth map of the previous video frame. Figure \ref{fig:occlusion_mask_disp_previous_frame} show the depth map prediction of the previous frame. Figure \ref{fig:occlusion_mask_sampled_disp_previous_frame} shows the depths that are sampled from the prediction \ref{fig:occlusion_mask_disp_previous_frame} using the sample locations from \ref{fig:occlusion_mask_grid_project_to_previous_no_depth}. By comparing the expected depths at the sample locations \ref{fig:occlusion_mask_expected_sample_depth} with the observed depths, i.e.  those sampled from the adjacent frame depth map prediction \ref{fig:occlusion_mask_sampled_disp_previous_frame}, the occluded regions can be determined \ref{fig:occlusion_mask_mask}. When the sampled observed depth is closer than what is expected from reprojection, it means something is blocking the view (i.e. occlusion). This occlusion mask can be used to ignore these regions in the reconstruction \ref{fig:occlusion_mask_proj_from_previous}.

Equation \ref{eq:occlusion_mask} shows the definition of the occlusion mask, which  indicates which of the pixels of the target image are visible in the image from which is sampled and thus can be used to ignore the occluded pixels in the reconstruction. In this formula $z_{t'*}$ is the value of the depth map prediction of frame $I_{t'}$ sampled at location $(x_{t \to t'}, y_{t\to t'})$, i.e. the location after the projection. 

If due to variation in the depth map predictions, an object is predicted closer in the adjacent frame, this could introduce regions that are incorrectly recognized as occlusion. To avoid this we propose to add the parameter $tolerance$ so that only regions with a sufficiently high ratio of predicted and expected (i.e. projected) depth are considered occluded. In our experiments we used $tolerance=0.3$ since this value produced occlusion mask that were good enough for performing our experiments. See section \ref{occlusion_mask_parameter_value_appendix} of the appendix for further considerations on choosing the parameter value.

The second case of equation \ref{eq:occlusion_mask} ignores the occlusion that is due to projected coordinates being outside of the image boundaries, just like the principled mask in  \cite{mahjourian_unsupervised_2018}.

\begin{equation}
\label{eq:occlusion_mask}
\omega_{t \to t'}  = \begin{cases} 0 & \text{if } z_{t'*} > z_{t \to t'} * (1 - tolerance) \\ 0 & \text{if } x_{t \to t'} \text{ or } y_{t \to t'} \text{ outside of image}\\ 1 & \text{otherwise} \end{cases}
\end{equation}
This computation of the occlusion mask does not introduce any new learnable parameters. Furthermore the adjacent frames depth map predictions are only used for determining the occlusion mask and not for backpropagation, therefore no gradients have to be computed which limits the computational overhead and thus the additional training time.

\subsubsection{Non-occluded average loss} 
\label{owa_loss}
Equation \ref{eq:owa_loss} shows the non-occluded average loss, a photometric loss function that uses the occlusion mask to average the reconstruction errors. The resulting loss map consists of the pixel wise average of the non-occluded regions of the reconstructions. For example, if a region is visible in both of the adjacent images, the loss for that region will be the average of both their reconstruction errors. If a region is only visible in one of the adjacent images, only the reconstruction error of that image is used. When a region is somehow occluded in both the adjacent images, none of the reconstruction errors will be used for training the model.
\begin{equation}
\label{eq:owa_loss}
L_p = \frac{\displaystyle\sum_{t'} \omega_{t \to t'} pe(I_t, I_{t' \to t})}{\max(\displaystyle\sum_{t'} \omega_{t \to t'}, 1)}
\end{equation}

\subsubsection{Non-occluded minimum reprojection loss}
\label{nomr_loss}
A photometric loss function that more resembles the one used in the baseline can be seen in equation \ref{eq:nomr_loss}. This loss function has the same properties as the per-pixel minimum reprojection loss, however it also incorporates the occlusion mask as a penalty. Per-pixel minimum reprojection loss uses only one reconstruction per pixel location in the loss. It is not known whether selecting only one reconstruction in a non-occluded region has an effect on the trained model. It is possible that this effect is beneficial to the model performance. If this were the case, averaging the non-occluded regions as done in equation \ref{eq:owa_loss} might harm performance. We will show experimentally that this is indeed the case: it helps to consider the per-pixel minimum also for non-occluded regions.

In equation \ref{eq:nomr_loss} you can see the photometric loss function that has the same behavior as equation \ref{eq:ppmr_loss} for non-occluded regions while further preventing occluded pixels from being used in the loss function, by using the mask as an additional error term. The implementation of the photometric error function has a convenient range of $[0,1]$, which makes scaling the occlusion mask unnecessary. For the model variant that uses this photometric loss, the occlusion mask is used in the same manner in the automask computation, i.e. as an additional error term on the right hand side of equation \ref{eq:automask}. This is done in order to prevent that areas that are occluded in both adjacent images, will be used to update the model. 
\begin{equation}
\label{eq:nomr_loss}
L_p = \min_{t'} (pe(I_t, I_{t'\to t}) + (1 -\omega_{t \to t'}))
\end{equation}
\section{Experiments}
We will evaluate the performance of models trained using the introduced photometric losses and compare them with the baseline. Additionally we will visualize the effect of the photometric loss function on the learning signal during training using examples from the training set.

\subsection{KITTI dataset}
In our experiment we will use the KITTI dataset since it is a widely used dataset in single-view depth estimation literature. This dataset contains short videos that are captured while driving through and around Karlsruhe in a car equipped with multiple cameras, a laser scanner and a GPS system.\cite{geiger_are_2012} We will use the split of this dataset introduced in \cite{eigen_depth_2014} with the static frames filtered out as described in \cite{zhou_unsupervised_2017}. This split uses 33 drives (the short videos) to produce 39,810 triplets for training and 4,424 for validation, and for testing 697 images sampled balanced from 28 other drives.

\subsubsection{Results}

The results in table \ref{tab:kitti_results} show that incorporating the occlusion mask into the used loss function improves the accuracy of depth prediction models. The model trained with non-occluded minimum reprojection loss outperforms the baseline, i.e. the per-pixel minimum reprojection loss model, on all metrics.  Incorporating the occlusion mask gives a larger accuracy improvement for the average reprojection loss function (table \ref{tab:kitti_results}a) than for the per-pixel minimum reprojection loss function (table \ref{tab:kitti_results}b). This can be explained by the fact that per-pixel minimum reprojection loss already ignores occlusion effects based on reconstruction error, and that further improvement can only come from ignoring occlusion effects that are missed by per-pixel minimum reprojection loss but are recognized by the occlusion mask.

\definecolor{light-gray}{gray}{0.85}
\newcommand{\grayline}{\arrayrulecolor{light-gray}\cline{2-10}\arrayrulecolor{black}}

\begin{table*}[htbp]
  \centering
  \resizebox{\textwidth}{!}
{
    \footnotesize
    \begin{tabular}{|l|l||c||c|c|c|c||c|c|c|}
      \hline
      &\multirow{2}{*}{Photometric loss} & \multirow{2}{*}{\makecell{Occlusion\\mask}} & \multicolumn{4}{|c|}{\makecell{Error metrics\\(lower  is better)}} & \multicolumn{3}{|c|}{\makecell{Accuracy metrics\\(higher is better)}}\\
      \cline{4-10}
      & & &
      Abs Rel & Sq Rel & RMSE  &
      RMSE$_{log}$& 
      $\delta < 1.25$ & $\delta < 1.25^{2}$ & $\delta < 1.25^{3}$ \\
      
      \hline

    (a) & Average reprojection* &   &
      0.117 &   0.878 &   4.846 &   0.196 &   0.870 &   0.957 &   0.980 \\ %
      \grayline
      & Non-occluded average reprojection & \checkmark & 0.117  &   0.936  &   4.908  &   0.195  &   0.874  &   0.958  &   0.980  \\
      
      \hline
          (b)  &  Per pixel min~reprojection \cite{godard_digging_2018}          & &
        0.114  &   0.915  &   4.874  &   0.193  &   0.877  &   0.959  &   {\bf   0.981}  \\
            
       \grayline
      & Non-occluded min reprojection & \checkmark &  {\bf 0.113 }&   {\bf   0.865  }&   {\bf   4.789 }&   {\bf   0.192  }&   {\bf   0.878 }&   {\bf   0.960 }&   {\bf   0.981 } \\
      
      \hline %
      
      \end{tabular}
  }
    \vspace{0pt}
      \caption{Results of the models trained with different photometric losses on the KITTI dataset \cite{geiger_are_2012}. Best results are in \textbf{bold}. Training is done using the subset of the Eigen split \cite{eigen_depth_2014} introduced for monocular training by \cite{zhou_unsupervised_2017}.
      *the numbers shown for this model are from table 2 in \cite{godard_digging_2018}.  $\delta= \max(\frac{d_{pred}}{d_{gt}}, \frac{d_{gt}}{d_{pred}})$
      \textbf{(a)} photometric losses that are based on averaging the photometric errors of both reconstructions. \textbf{(b)} photometric losses that are based on selecting the reconstruction with the lowest reconstruction error. 
      Both types of loss functions (a, b) get improved result when occlusion mask is incorporated.
      }
      \vspace{-2pt}
\label{tab:kitti_results}
\end{table*}

Although the observed performance improvements validate the benefit of incorporating occlusion mask into the loss function, the results do not validate the hypothesis, which states that more useful information is available for model training if only those regions that suffer from occlusion are ignored. Given this hypothesis, it would be expected that the non-occluded average reprojection model outperforms the per-pixel minimum reprojection model.

We can think of various explanations for this result: the occlusion mask fails to mask occluded regions, the occlusion mask incorrectly masks non-occluded regions, or per-pixel minimum reprojection loss ignores regions which are not occluded but that do have a negative effect on training the model. We will try to answer this question in the next section by analyzing the loss images created by the discussed photometric losses.

\FloatBarrier

\subsubsection{Visualizing Photometric Losses}

In this section we will compare loss images of the two proposed photometric losses (i.e. the non-occluded average reprojection loss from section \ref{owa_loss} and the non-occluded minimum reprojection loss from section \ref{nomr_loss}) with loss images created by the baselines photometric loss (i.e. per pixel minimum reprojection loss, section \ref{ppmr_loss}). These loss images are computed on training examples and provide an idea of how important the regions of a reconstructed image are for optimization during training.  

We intend to identify the differences between the loss functions. For example if one of the methods fails to ignore occluded regions or unexpectedly ignores non-occluded regions. Additionally this analysis can demonstrate whether the occlusion mask is able to mask occlusion. Since the loss function is used to guide the learning behavior during training, we will apply our analysis on the data from the training set. The loss images are all created using depth predictions made by the same model, i.e. one trained with per-pixel minimum reprojection loss. We assume that model choice has limited influence on the results of this analysis. Training examples for visualization will be selected randomly as well as based on the computed loss values. The criterion for selection is a large absolute difference between the computed per-pixel minimum reprojection loss and the non-occluded average reprojection loss. To avoid having many similar examples, we filter the data by only looking at images from one camera and at every tenth frame recorded.

\begin{figure*}[htbp]
    \centering
    \includegraphics[width=\textwidth]{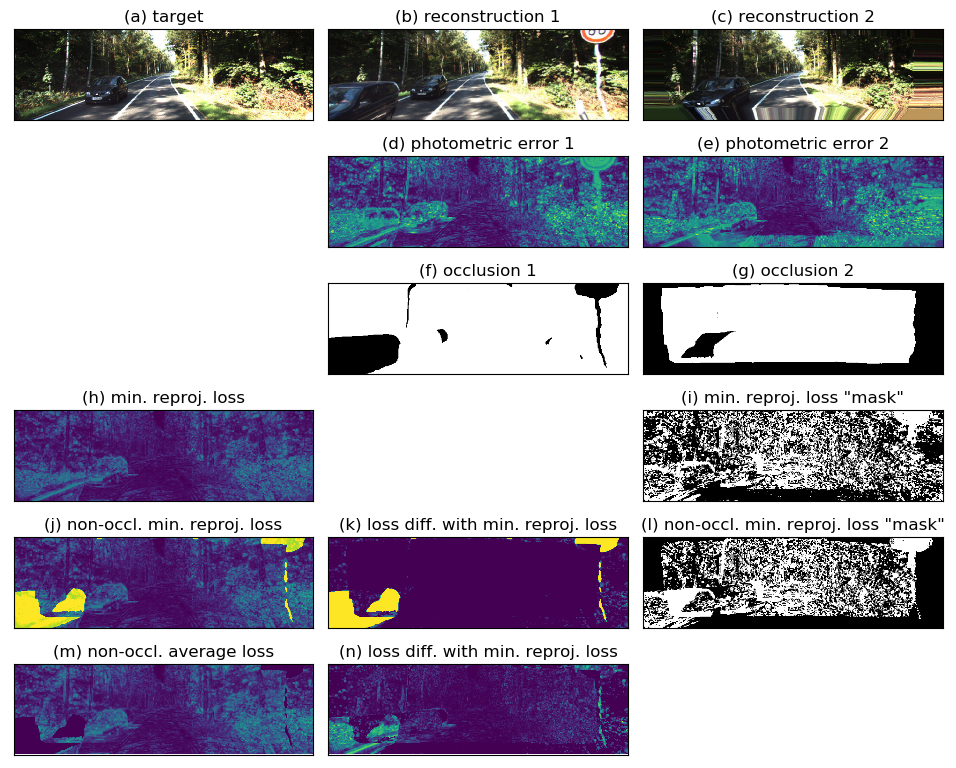}
    \caption{\textbf{Visualizing photometric losses:}
    Example with the largest difference between between the per-pixel minimum reprojection loss and the non-occluded  average reprojection loss. 
    \textbf{(a)} The reconstruction target.
    \textbf{(b\&c)} Reconstructions made from video frames adjacent to the target.
    \textbf{(d\&e)} Photometric errors from (b\&c) to (a).
    \textbf{(f\&g)} Occlusion masks depicting which areas are visible of (b\&d) and (c\&e) respectively.
    \textbf{h, j\& m:} Visualisations of the computed per-pixel minimum reprojection loss, non-occluded minimum reprojection loss, and non-occluded average reprojection loss, respectively. 
    \textbf{(k\&n)} Absolute differences of (j\&m) with (h)
    \textbf{(i\&l)} Masks displaying for each pixel location which reconstruction's photometric error is used in the loss (black =1, white=2) by (h) and (j), respectively.
    } 
    \label{fig:loss_diff_highest}
\end{figure*}

Figure \ref{fig:loss_diff_highest} shows the example with the highest difference in the computed loss. 
The first five images of this figure show the target frame, both reconstructions, and the photometric errors of the reconstructions. In the photometric error images, a brighter color means a higher value. The images of the third row show the calculated binary occlusion masks, where white areas mean there is no occlusion and black means there is occlusion. The last three rows show for each photometric loss, the calculated loss maps, the absolute difference to the baseline photometric loss and if applicable a mask displaying for each pixel location which of the photometric errors is used in the loss (black means from reconstruction 1, white from reconstruction 2). An important remark is that the shown loss maps are the photometric loss where the automask has not yet been applied. 

This example has a high absolute difference for the two loss methods, because there is an area that is occluded in both of the adjacent frames and per-pixel minimum reprojection loss is only able to ignore one of the reconstructions. This example shows that the occlusion mask (\ref{fig:loss_diff_highest}f+g) is indeed able to mask the occluded areas, i.e. the black regions of the occlusion masks match with regions that are visible in the target image but not in the reconstructions. The piece of road that is visible in the target frame is occluded by the first car in the previous frame (\ref{fig:loss_diff_highest}b) and the tailgating car in the later frame (\ref{fig:loss_diff_highest}c). Together these occlusions make it impossible to reconstruct that area of the target image. What the example also shows, is that the occlusion mask is able to handle thin objects, which can be seen by the traffic sign in \ref{fig:loss_diff_highest}b,d and f.

The areas that are occluded in both of the adjacent images are visible as the black regions in the non-occluded average loss image (\ref{fig:loss_diff_highest}m), because they are directly ignored using the occlusion mask. In the loss image of the non-occluded minimum reprojection loss (\ref{fig:loss_diff_highest}j) these areas are clearly visible as the yellow areas because the occlusion mask is added to the loss, and consequently will be ignored by the automask.

\begin{figure*}[htbp]
    \centering
    \includegraphics[width=\textwidth]{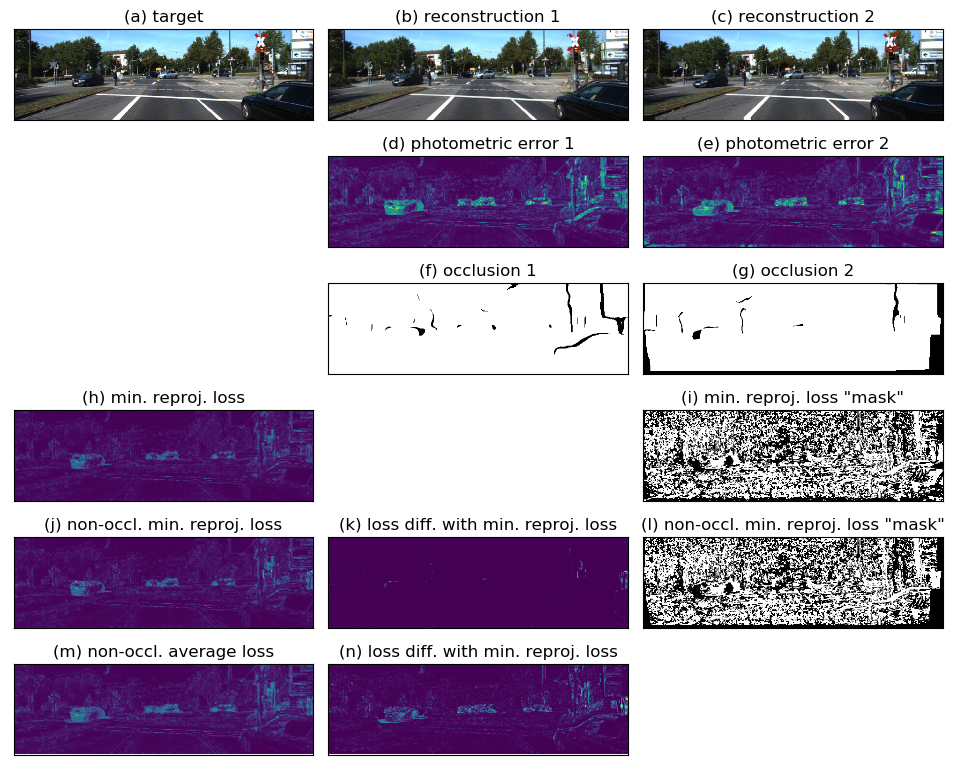}
    \caption{\textbf{Visualizing photometric losses:}
    Example that was randomly selected. In \textbf{(n)} we can see that the per-pixel minimum reprojection loss is able to ignore some of the reprojection error caused by object motion.
    }
    \label{fig:loss_diff_random1}
\end{figure*}

Figure \ref{fig:loss_diff_random1} shows one of the examples that was selected randomly. We can observe that the per-pixel minimum reprojection loss is able to ignore some of the reprojection error that is caused by object motion instead of occlusion. In the image which shows the loss difference between the non-occluded average loss and the minimum reprojection loss (fig\ref{fig:loss_diff_random1}n), we can see that there is a loss difference for moving objects, in this case the cars, which are not occluded. Although this behavior is not the originally intended effects of the per-pixel minimum reprojection loss it does help training, since the current reprojection computation (eq. \ref{eq:sampling}) does not account for object movement, and consequently any gradient coming from a moving object can interfere with learning correct depth predictions.

\section{Discussion}
The goal of this thesis was to design an occlusion mask that can be used to specifically discard regions that cannot be reconstructed due to occlusions and to test whether this method can help improve the performance of depth prediction models. We hypothesized that by limiting the amount of information which is ignored during training to only those regions that suffer from occlusion, more useful information will be available for training the model. 

Here, we have introduced an occlusion mask that is based entirely on depth predictions and can be used to specifically ignore regions where occlusion is expected. We have shown that incorporating the occlusion mask in the photometric loss function improves model performance. 

Contrary to our hypothesis, the results of our experiments show that ignoring only regions where occlusion is expected (non-occluded average loss) did not give better results compared with per-pixel minimum reprojection loss. We further investigated this outcome by performing a visual analysis of the difference between the photometric losses obtained by both methods.  

In the visual analysis of the photometric losses, we observed (in fig. \ref{fig:loss_diff_highest}) that with the occlusion mask it is possible to ignore areas of reconstructions that are occluded in both of the adjacent frames. This is an improvement compared to the per-pixel minimum reprojection loss which is only able to ignore one of the reconstructions. 

During this analysis we have also discovered (in fig. \ref{fig:loss_diff_random1}n) that the per-pixel minimum reprojection loss unexpectedly reduces the loss being contaminated with photometric error caused by object motion. This effect takes place because the per-pixel minimum reprojection loss always ignores one of the reconstructions at each pixel location based entirely on a high photometric error, which itself is not caused exclusively by occlusion. The ability to partially ignore moving objects is useful for models that do not take object motion into account.

It makes sense that a model which does not take object motion into account does not benefit from incorporating errors caused by object motion into the loss function. This however does not give a complete answer about the reason for the observed performance gap between the non-occluded average photometric loss and the per-pixel minimum reprojection loss. It is possible that for the per-pixel minimum reprojection loss the improvement in model performance which is gained by ignoring motion artefacts is  at the same time reduced by ignoring regions which are not affected by occlusion or motion and do have a valid gradient.

The question of whether ignoring only the areas of reconstructions that are affected by occlusion during training gives a better performing model, remains for models that do take object motion into account. 

It is possible that (partially) ignoring object-motion related photometric error, as is done by the per-pixel minimum reprojection loss, could be disadvantageous for models that do take object motion into consideration since it allows the model to achieve a lower loss while making incorrect predictions about object movement. Occlusion mask does not have this problem because it can specifically ignore occlusion effects, while preserving the photometric error caused by object movement which is valuable information when optimizing the model.

Looking back to the implicit assumptions underlying self-supervised training using image reconstruction\footnote{\begin{enumerate}[noitemsep]
\item the scene is static without moving objects; 
\item there is no occlusion/disocclusion between the target view and the source views;
\item the surfaces appear uniformly bright from all directions of view so that the photo-consistency error is meaningful
\end{enumerate}
} that were mentioned in section \ref{image_reconstruction}, and combining them with our earlier observations we can see that the per-pixel minimum reprojection loss is able to ignore some or most of the photometric error related to the first two assumptions, i.e. the scene is static and there is no occlusion. We can imagine that per-pixel minimum reprojection loss is also able to ignore some of the photometric error related to the third assumption, i.e. surface appearance changes with direction of view. Or changes in surface appearance caused by a change in illumination due to moving objects, e.g. shade or reflection.

\section{Conclusion}
In this work we have introduced occlusion mask, a mask that during training can be used to specifically ignore regions that cannot be reconstructed due to occlusions. \mbox{Occlusion} mask is based entirely on predicted depth information. We have demonstrated that (i) incorporating occlusion mask in the used photometric loss function can improve the performance of single image depth prediction models. (ii) per-pixel minimum reprojection loss also ignores some of the reprojection error caused by object motion.

\FloatBarrier

\bibliographystyle{ieee}
\bibliography{references}

\clearpage

\begin{appendices}
\section{Occlusion mask - parameter value}
\label{occlusion_mask_parameter_value_appendix}
In this work a “tolerance” parameter is introduced in the implementation of the occlusion mask to prevent incorrectly recognizing regions as occluded due to variation in the predicted depth maps of adjacent frames. Due to time constraints limited consideration has been put in choosing the value that was used. 

We think that the “optimal” value depends on the performance of the model for which it is used. For example if a model gives almost perfect depth map predictions, the value could be (very close) to zero. Another example would be a model that gives more accurate depth predictions for closer objects than for objects further away. In this situation the “tolerance” value can be based on the distance to the object. One of the reasons for not further investigating these options is that considering that the choice of parameter value is dependent on the network performance, it means that when future models keep improving, the value choice becomes less important and just taking a constant value close to zero will work just fine. 

During the design of the occlusion mask when determining what value to use for the“tolerance” parameter, we forgot to put the network into evaluation mode, which resulted in the model producing less accurate depth maps. This could have resulted in overestimating the depth map prediction variability and thus choosing an unnecessarily high tolerance value. A consequence of a high value for this parameter are that some occlusions are not recognized when “the occluded” and “the occluding” are close, and the ratio of the distances to them is small. An example of this can be seen in figure \ref{fig:loss_diff_random1} where “the occluding” parts of the moving car that are close to the road, are not marked as occluding in the occlusion mask and are visible in the non-occluded average loss image.

\end{appendices}

\end{document}